\pdfoutput=1
\documentclass[11pt]{article}

\usepackage[preprint]{acl}

\usepackage{times}
\usepackage{latexsym}

\usepackage[T1]{fontenc}
\usepackage[utf8]{inputenc}

\usepackage{microtype}

\usepackage{inconsolata}

\usepackage{graphicx}

\usepackage{lipsum} 
\usepackage{booktabs}
\usepackage{multirow}
\usepackage{graphicx}
\usepackage[normalem]{ulem}
\useunder{\uline}{\ul}{}
\usepackage{amsmath}
\usepackage{enumitem}
\usepackage{subcaption}
\usepackage{array}
\usepackage{tcolorbox}

\PassOptionsToPackage{prologue,dvipsnames}{xcolor}
\usepackage[dvipsnames]{xcolor}

\usepackage{cleveref}

\usepackage{xspace}
\newcommand{\ours}{\textsc{AD-Agent}\xspace}

\title{\ours: A Multi-agent Framework for End-to-end Anomaly Detection}

\author{
    \textbf{Tiankai Yang\textsuperscript{1}$^{,*}$},
    \textbf{Junjun Liu\textsuperscript{1}$^{,*}$},
    \textbf{Wingchun Siu\textsuperscript{1}$^{,*}$},
    \textbf{Jiahang Wang\textsuperscript{1}},
\\
    \textbf{Zhuangzhuang Qian\textsuperscript{1}},
    \textbf{Chanjuan Song\textsuperscript{1}},
    \textbf{Cheng Cheng\textsuperscript{2}},
    \textbf{Xiyang Hu\textsuperscript{3}},
    \textbf{Yue Zhao\textsuperscript{1}}
\\
    \textsuperscript{1}University of Southern California, 
    \textsuperscript{2}Carnegie Mellon University,
    \textsuperscript{3}Arizona State University
\\
    \texttt{
        \{tiankaiy, junjunl, siuw, jwang857, alexqian, songchan\}@usc.edu,
    }
    \\
    \texttt{
        ccheng2x@gmail.com, xiyanghu@asu.edu, yzhao010@usc.edu
    }
\\
    \vspace{-9pt}
}
\newcommand\nnfootnote[1]{%
  \begin{NoHyper}
  \renewcommand\thefootnote{}\footnote{#1}%
  \addtocounter{footnote}{-1}%
  \end{NoHyper}
}

\definecolor{keygreen}{HTML}{008000}

\begin{document}
\maketitle
\begin{abstract}
% \vspace{-3pt}
% \tk{short version}
Anomaly detection (AD) is essential in areas such as fraud detection, network monitoring, and scientific research. 
However, the diversity of data modalities and the increasing number of specialized AD libraries pose challenges for non-expert users who lack in-depth library-specific knowledge and advanced programming skills.
To tackle this, we present \ours, an LLM-driven multi-agent framework that turns natural-language instructions into fully executable AD pipelines.
\ours coordinates specialized agents for intent parsing, data preparation, library and model selection, documentation mining, and iterative code generation and debugging. Using a shared short-term workspace and a long-term cache, the agents integrate popular AD libraries like PyOD, PyGOD, and TSLib into a unified workflow.
Experiments demonstrate that \ours produces reliable scripts and recommends competitive models across libraries. 
The system is open-sourced to support further research and practical applications in AD.

% \tk{long version}
% Anomaly detection (AD) is essential in areas such as fraud detection, network monitoring, and scientific research.
% However, with the diversity of data modalities and the increasing number of specialized AD libraries, building reliable AD workflows typically requires in-depth knowledge of multiple incompatible libraries and non-trivial programming effort. This poses challenges for non-expert users in quickly testing ideas.
% To tackle this, we present \ours, an LLM-driven multi-agent framework that turns natural-language instructions into fully executable AD pipelines.
% \ours coordinates specialized agents for intent parsing, data preparation, library and model selection, documentation mining, and iterative code generation and debugging. Using a shared short-term workspace and a long-term cache, the agents integrate popular libraries like PyOD, PyGOD, and TSLib into a unified workflow.
% On standard AD benchmarks, \ours automatically produces runnable scripts in over 80\% cases and recommends competitive models across libraries.
% The system is open-sourced to support further research and practical applications in the AD community.
\end{abstract}

\newcommand{\gitlink}{https://github.com/USC-FORTIS/AD-AGENT}
\nnfootnote{$^*$Equal contribution.}

\vspace{-15pt}
\section{Introduction and Related Work}
\vspace{-3pt}

% With the continuous growth of anomaly detection, the community has developed a bunch of well-designed libraries for various data types

% \tk{need to reduce repetitive content about introducing \ours.}
% \tk{add a shortcut of conversation here, a dialogue? I will add a complete conversation in the appendix.}

Anomaly detection (AD) plays a crucial role in a wide range of applications, including fraud detection~\citep{abdallah2016fraud}, network monitoring~\citep{sun2023efficient}, action recognition~\citep{li2024dpu}, and medical analysis~\citep{fernando2021medical}. 
% As the field matures, the community has developed specialized open-source libraries to support different data modalities, each offering a diverse collection of robust models tailored to its domain. 
To handle these diverse data types, the community has released modality-specific open-source libraries that package state-of-the-art models and utilities.
% While these resources have accelerated research and deployment, they also demand technical familiarity with incompatible library-specific functions.
Although these libraries accelerate experimentation, each introduces its own data formats and APIs, so users must ``juggle'' incompatible workflows before they can run even baseline methods. 
This learning overhead discourages adoption, especially among domain specialists who are not software/data engineers.
% This raises a significant barrier for both researchers and practitioners, particularly those without extensive programming or domain expertise. 
{The stakes are high: 
Knight Capital lost USD 440 million in 45 minutes when an unchecked trading anomaly cascaded through its systems \citep{KnightCapital2012}, 
and Target’s 2013 breach
% , intensified by dismissed intrusion alerts, 
has cost more than 200 million 
\citep{TargetBreachKillChain}. 
These incidents show that small gaps in an AD pipeline can cause major financial or security failures, showing the need for tooling that is both reliable and easy to integrate.
}

Meanwhile, large language models (LLMs) have demonstrated strong capabilities in reasoning~\citep{guo2025deepseek}, code generation~\citep{liu2023code}, and tool use~\citep{schick2023toolformer}. 
Recent advances in agent-based systems have further enhanced the potential of LLMs to automate complex, multi-stage tasks that previously required substantial manual effort~\citep{guan2023leveraging} (see extended related work in Appx. \ref{sec:related}). 
This presents a compelling opportunity:
\textit{Can we develop a general-purpose AD platform that leverages LLMs and existing libraries to build complete detection pipelines from the natural language intents of non-expert users?}
% \textit{Can we build a general-purpose AD platform that leverages LLMs and existing libraries to build complete detection pipelines from high-level user instructions?}

% % We introduce \ours, 
% % We answer this question with \ours, 
% To address this, we introduce \ours, 
% an LLM-powered, collaborative multi-agent framework that turns natural-language instructions into end-to-end AD solutions across modalities.
% % a collaborative multi-agent framework powered by LLMs that unifies AD across modalities. 
% \ours decomposes the workflow into specialized agents that interpret user intent, process data, select libraries and algorithms, retrieve algorithmic knowledge, generate and verify code, and optionally evaluate and tune the pipeline.
% Short-term shared memory coordinates agent interaction, while a long-term cache eliminates redundant web queries for algorithm metadata.
%
% To address this, we introduce \ours --\yz{a multigent .... to...}. 
% It decomposes the AD workflow into specialized agents responsible for interpreting user intent, processing data, selecting libraries and models, retrieving algorithmic knowledge, generating and verifying code, and optionally evaluating and tuning the pipeline. 

\begin{figure}[!t]
\vspace{-5pt}
\centering
\setlength{\abovecaptionskip}{10pt}
\includegraphics[width=0.43\textwidth]{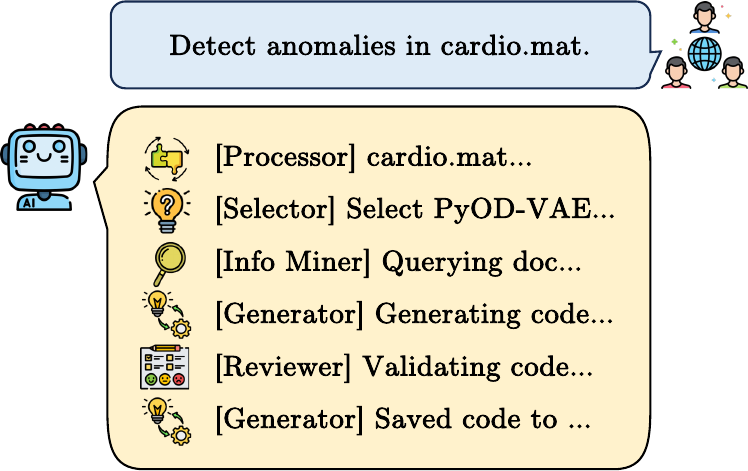}
\vspace{-5pt}
    \caption{Illustration of \ours: given a user request, the multi-agent system coordinates each stage to generate a runnable pipeline.
   }
   \label{fig:short_cut}
   \vspace{-20pt}
\end{figure}

\begin{figure*}[!t]
\vspace{-5pt}
    \centering
    \includegraphics[width=0.8\linewidth]{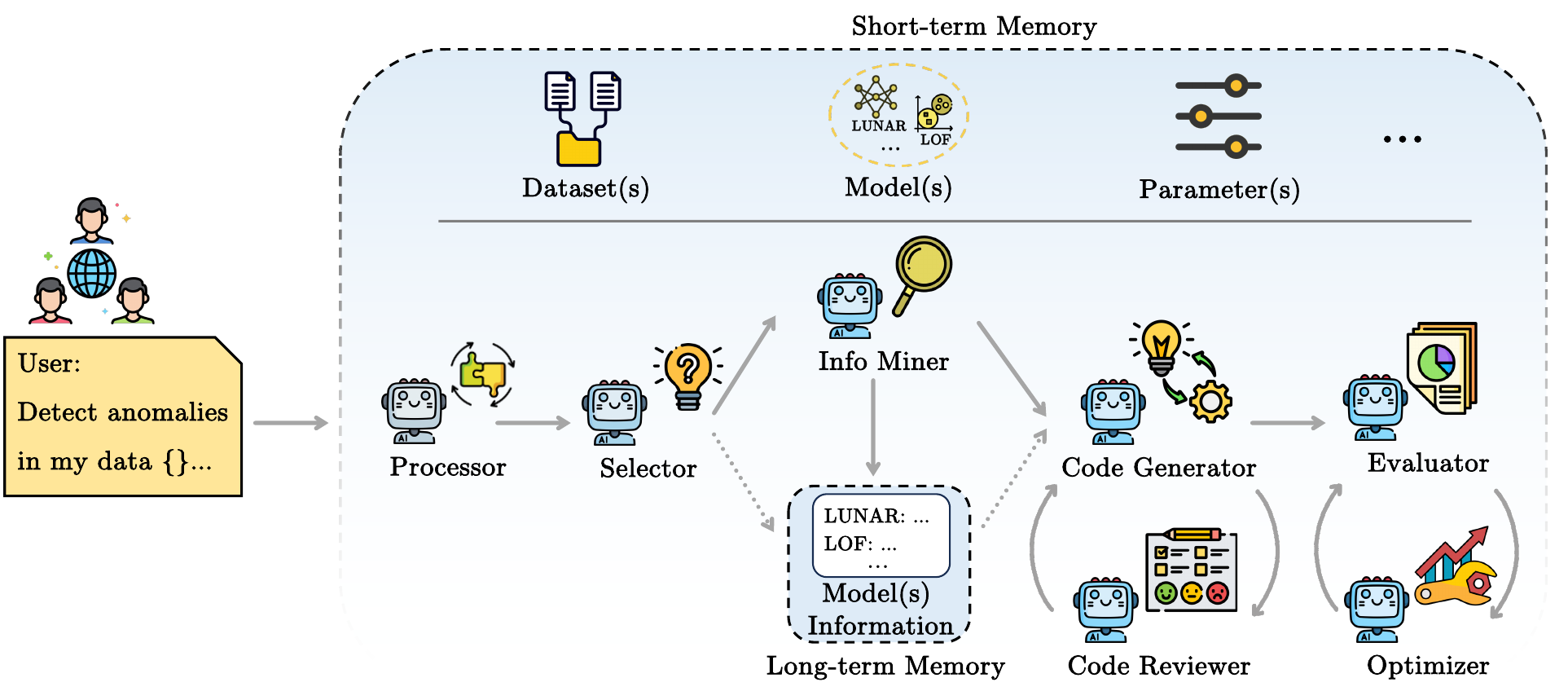}
    % \includegraphics[width=1\linewidth]{figures/flowchart.pdf}
    % \caption{Flowchart of \ours. Users input natural language instructions and data from various modalities, such as multivariate, graph, or time series. \ours coordinates multiple LLM-powered agents via short-term and long-term memory to construct anomaly detection pipelines utilizing libraries like PyOD, PyGOD, and Orion. Solid arrows represent the default workflow; dashed arrows indicate an optional path that bypasses web searches when algorithm information is stored in long-term memory.}
    \vspace{-12pt}
    \caption{Flowchart of \ours. Users input natural language instructions and data from various modalities. \ours coordinates multiple LLM-powered agents via short-term and long-term memory to construct anomaly detection pipelines. Solid arrows represent the default workflow; dashed arrows indicate an optional path that bypasses web searches when algorithm information is stored in long-term memory.}
    \vspace{-18pt}
    \label{fig:overview}
    % \vspace{-18pt}
\end{figure*}

To address this, we introduce \ours -- a multigent framework powered by LLMs that automates the construction of AD pipelines from plain language instructions.
It decomposes the AD workflow into specialized agents responsible for user intent interpretation, data processing, library and model selection, knowledge retrieval, code generation and verification, and optional evaluation and tuning.
For the memory mechanism, which is the key component to support agent-environment interactions~\citep{memory_survey}, we propose two memories. The short-term shared memory maintains the context of the current session, enabling coordination among agents, while the long-term memory serves as a cache to reduce costly queries across repeated sessions.
% \yz{Why do we need short and long memory -- need a motivation in the last paragraph or call it out here. Also, any similar design short/long memory paper we could cite?}
% Short-term shared memory coordinates agent interactions, while a long-term cache relieves redundant web queries for model metadata.
% In this way, \ours allows non-expert users to create comprehensive AD pipelines across multiple libraries and modalities using natural language instructions, 
By combining specialized agents with structured memory, \ours allows non-expert users to build comprehensive AD pipelines across multiple libraries and modalities using only natural language, relieving the need for library-specific expertise or manual programming.
Figure~\ref{fig:short_cut} provides an illustration of \ours.
% We introduce \ours, a collaborative multi-agent framework that bridges high-level user intent and low-level implementation details. Powered by LLMs, our system decomposes the anomaly detection workflow into specialized agents that select appropriate tools, prepare data, mine implementation knowledge, generate executable code, and interpret results. 
% By combining the modularity of agent-based systems with the reasoning and synthesis capabilities of LLMs, \ours allows users to interact with anomaly detection workflows through natural language while benefiting from the breadth and depth of existing libraries.

% \noindent
A survey of prior related LLM-agent work and modality-specific AD libraries is provided in Appendix~\ref{sec:related}.
\textbf{Our contributions are as follows}:
\begin{itemize}[nosep, itemsep=3pt, leftmargin=*]
    % \item \textbf{Unified LLM-based automation}.
    % \item \textbf{Unified multi-modality, multi-library automation}.
    \item \textbf{Unified multi-modal-library automation}.
    We propose the first multi-agent framework that integrates multiple domain-specific AD libraries, enabling end-to-end, cross-modality pipeline construction from natural language.
    % \item \textbf{Unified multi-modality and multi-library automation}. We propose the first multi-agent framework that supports a broad range of data modalities and integrates multiple domain-specific AD libraries.
    
    % that integrates multiple domain-specific AD libraries, enabling end-to-end, cross-modality pipeline construction from natural language.
    
    % \item \textbf{Modular, extensible, and long-lifecycle design}. \ours is built from loosely coupled agents for reasoning, retrieval, and generation. This design allows for easy adaptation to new libraries and tasks with minimal modifications as the ecosystem evolves.
    \item \textbf{Modular, extensible, and long-lifecycle design}. Loosely coupled agents for reasoning, retrieval, and generation enable \ours to easily incorporate new libraries and tasks with minimal changes, supporting a long-lasting ecosystem.
    \item \textbf{Accessible to non-experts}. \ours converts natural language instructions into executable scripts and supports diverse data types, enabling non-expert users without programming skills or specialized knowledge to start easily.
    \item \textbf{Open-source release}. We release \ours at \ \ \url{\gitlink} to provide the community with a practical, extensible platform for LLM-driven AD research and real-world applications. 
    % \yz{replace to anonymous link}
    % \tk{Will create one once we double check there is no personal information in the github repo tomorrow}
    % \item We demonstrate that \ours handles diverse real-world scenarios, supporting non-expert users through natural language input and flexible data formats.
    % \item \yz{I believe we should mention the support of multiple modalities and libraries somewhere as a highlight}
\end{itemize}

% \input{sections/1_related}
% \vspace{15pt}
\section{Methodology}
\vspace{-5pt}
% We proposed \ours, a versatile, integrated, automated end-to-end multi-agent framework for the anomaly detection pipeline. 

% In this section, we first give an overview of our proposed multi-agent framework \ours. Next, we introduce the task decomposition and their corresponding agents. Finally, we discuss the collaboration between multiple agents.

% \subsection{Overview}
We present \ours, a multi-agent framework that automates AD across diverse modalities and use cases. By integrating established AD libraries -- PyOD for multivariate data, PyGOD for graph data, and TSLib for time series -- \ours supports a broad range of models and enables end-to-end automation from user instruction to script.

% A specialized agent powered by LLMs manages each stage of the workflow. These agents collaborate via a centralized working memory to interpret user intent, preprocess data, retrieve algorithmic knowledge, generate runnable code, and evaluate results. By combining modular agent design with the reasoning capabilities of LLMs, \ours adapts flexibly to different tasks and user requirements while minimizing the need for manual programming or domain-specific expertise.

% We propose \ours, a collaborative multi-agent framework that automates anomaly detection across diverse data modalities. By integrating with established anomaly detection libraries—PyOD for tabular data, PyGOD for graph data, and TODS for time series—ADAgent enables end-to-end automation from user instruction to executable code. Agents specialize in different subtasks and coordinate via a centralized working memory, referred to as the. This modular structure allows the system to adapt to various data types and user demands while minimizing the need for manual programming or domain-specific expertise.
\vspace{-5pt}
\subsection{Agents}
\vspace{-3pt}
We decompose the detection workflow into multiple subtasks, with each stage handled by a specialized LLM-powered agent, as illustrated in Fig.~\ref{fig:overview}.

\noindent\textbf{Processor}. Datasets in practice come in diverse formats (e.g., .csv, .mat, or even natural language), and detection tasks may vary from supervised setups to zero-shot scenarios. 
The Processor agent serves as the entry point of the system, using LLMs to interpret inputs, infer key attributes (e.g., modality, supervision type), and extract user-specified constraints. It organizes this information into a structured format that guides downstream agents.
% The Processor agent leverages LLMs to interpret the input, infer key attributes like data modality, structure, and supervision type, and extract any user-stated preferences or constraints. This information is organized into a structured representation that guides downstream agents in library selection and pipeline construction.

% The Processor agent uses LLMs to parse the data and infer the task context. 
% \tk{revise this part: }
% It restructures inputs into forms compatible with the chosen library and extracts detailed user constraints, such as desired algorithms or parameter settings.

% \noindent\textbf{Selector} agent analyzes both the data and the accompanying prompt to determine its modality and choose the appropriate anomaly detection library. Rather than relying on strict rules, the agent utilizes the reasoning capabilities of LLMs to match the task with the correct library, even in challenging yet common situations where the input lacks a clear structure or format specification.

\noindent\textbf{Selector}. Building on the Processor’s output, the Selector agent determines which AD library best aligns with the inferred data modality and task requirements. 
% Rather than relying on rigid heuristics, it uses LLM-based reasoning to match the task with the correct library, even in challenging yet common situations where the input lacks a clear structure or format specification. 
If the user does not specify a model, the Selector recommends one from the chosen library. Inspired by recent advances in LLM-based model selection~\citep{qin2025metaood, chen2024pyod, adllm}, it leverages the LLM’s knowledge of models to provide context-aware suggestions tailored to the dataset and task. 
% his approach relieves the need for domain expertise. 
% See details in Appx.~\ref{}.

% If the user does not specify an algorithm, the Selector further recommends a suitable model from the chosen library. Instead of following static rules, the agent leverages the LLM’s broad knowledge of algorithmic behavior, design assumptions, and practical guidelines. This allows it to make informed suggestions based on dataset characteristics and task context, providing flexible and context-aware model selection without requiring domain expertise from the user.

% uses LLM-based reasoning to handle ambiguity in the input and choose an appropriate library—PyOD, PyGOD, or TODS—even in the absence of explicit user guidance.

% \noindent\textbf{Processor}. In real-world applications, users usually have datasets in various file formats (e.g., ``.csv", ``.mat," or even the natural language description) and different detection scenarios (e.g., with training set or zero-shot detection). This requires the users to have background knowledge of both data processing in programming language and the usage of the corresponding library. Thanks to LLMs' strong understanding capability, \textbf{Provessor} agent reads and transfers the input into proper formats aligned with the requirement of the selected library automatically. It also analyzes users' requirements on algorithm selections and more details, such as the parameters' values.

\noindent\textbf{Info Miner}. Understanding how to apply a model often requires consulting multiple documentation sources, which can be time-consuming and challenging, especially for non-experts. The Info Miner agent performs this background research autonomously. It integrates ``Web Search" function from OpenAI~\citep{openaiwebsearch} to learn from and summarize relevant documents, code examples, and online tutorials. The output includes model descriptions, instructions, and parameter definitions for later code generation. 
% See details in Appx.~\ref{}.

% \noindent\textbf{Info Miner}. In traditional anomaly detection jobs, users need to read official documents, examples, tutorials on the Internet, and even the source code to master the usage of the selected library. It requires a large amount of time and effort and raises great challenges for non-expert users. To solve these issues. \textbf{Info Miner} integrates Web Search from OpenAI to read, learn, and extract the information in official documents, examples, online tutorials, and even the source code. Especially it summarizes a structured message, including algorithm descriptions, parameter definitions, initialization signatures, and example usages for later usage

% \noindent\textbf{Code Generator}. This agent translates planning into execution. Using the user instructions and knowledge from the Info Miner, it composes runnable Python code tailored to the selected algorithm. The code respects the data format, chosen parameters, and conventions of the selected library.

% \noindent\textbf{Code Reviewer}. LLMs can produce syntactically correct but semantically flawed code. The Code Reviewer Agent acts as a safeguard. It performs dry runs, identifies issues such as missing imports or invalid parameter names, and repairs them when possible. This layer of validation ensures robustness before any actual execution.

\noindent\textbf{Code Generator \& Reviewer}. These two agents collaborate to produce reliable detection scripts. The Generator composes code based on user instructions and knowledge from the Info Miner. To ensure correctness, the Reviewer validates the code through a dry run using LLM-generated synthetic samples, aiming to quickly catch any execution errors. If issues are detected, the two agents enter a feedback loop, iteratively refining the code until a valid and executable pipeline is achieved.

% \noindent\textbf{Code Reviewer} agent analyzes the generated code for syntactic and semantic errors to avoid possible hallucinations from LLMs' generated code. It catches the errors raised in the code execution and reflects to fix and improve the code. It simulates a dry run to detect issues like missing imports or invalid parameters and applies minor repairs if necessary.

% \noindent\textbf{Evaluator}. The final step in the workflow is not just execution but reflection. The Evaluator Agent runs the pipeline and interprets results. It provides performance summaries and, when needed, recommends next steps, such as trying alternative algorithms or adjusting configurations.

\noindent\textbf{Evaluator \& Optimizer}. These two agents provide optional extensions for performance evaluation and hyperparameter tuning. The Evaluator runs the pipeline and summarizes detection results when ground truth labels are available for the target dataset. The Optimizer, inspired by \citet{liu2025agenthpo}, performs LLM-powered hyperparameter tuning based on the provided training dataset. They operate in a feedback loop, iterating between parameter updates and performance assessment.

\vspace{-5pt}
\subsection{Agent Collaboration and Workflow}
\vspace{-3pt}
\ours facilitates collaboration through two memory structures: a shared \textbf{short-term memory} and a persistent \textbf{long-term memory}.

The short-term memory serves as the central workspace where agents read and write task-related content. It stores the user input, the processed dataset, selected models, and parameter configurations. This enables agents to operate independently while remaining context-aware.

The long-term memory caches model information retrieved by Info Miner. Since mining from web sources is often time-consuming and resource-intensive, the system first checks this cache for recent summaries before initiating a new web search. It is refreshed periodically (e.g., weekly), allowing the system to benefit from up-to-date resources while avoiding redundant queries. 
% This significantly reduces latency and cost without compromising access to relevant knowledge.

As shown in Fig.~\ref{fig:overview}, the system begins with the Processor, which interprets the user’s input and prepares the data. Based on this context, the Selector determines the appropriate library and, if unspecified by the user, recommends a suitable model. The Info Miner then gathers relevant model details, consulting either the long-term memory or the web. With this knowledge in place, the Code Generator and Reviewer collaboratively assemble and verify the detection pipeline through an iterative feedback loop until the code is valid and executable. Users may then choose to enable the Evaluator and Optimizer for optional performance assessment and hyperparameter tuning. 

This collaborative agent framework allows \ours to flexibly support multiple data types, including new libraries, adapt to varying input formats, and deliver usable outputs with minimal user effort. Each agent contributes a specialized capability, with LLMs enabling reasoning, adaptation, and coordination across the workflow.

\vspace{-2pt}
\section{Experiments}
\vspace{-3pt}
We evaluate \ours on reliability and efficiency of the system in constructing executable anomaly detection pipelines from natural language instructions, the quality of model selection, and the effectiveness of long-term memory. 
% See more discussions in Appx.~\ref{sec:appx_use_case} about use cases and Appx.~\ref{sec:appx_optimizer} for Optimizer improvement.
For discussion of use cases, see Appx.~\ref{sec:appx_use_case}.
Improvements by Optimizer can be found in Appx.~\ref{sec:appx_optimizer}.

\noindent
\label{sec:exp_benchmark}
\textbf{Datasets and Models}. We select datasets and models for each library from their corresponding benchmarks: \citet{chen2024pyod} for PyOD, \citet{liu2022bond} for PyGOD, and \citet{tslib_bench} for TSLib. See details in Appx.~\ref{sec:appx_dataset_model}.

\begin{table}[t]
% \vspace{-0.3in}
\setlength{\abovecaptionskip}{3pt}
\caption{
% Pipeline generation performance by library. Success rate is the percentage of dataset–algorithm pairs whose generated code executed without errors. Time is the average end-to-end generation latency. In/Out Tokens reports the average number of input and output tokens consumed by the LLM. Cost reflects the estimated per-pipeline LLM billing in US dollars.
Pipeline generation performance by library, showing success rate (code runs without error), average latency, LLM token usage (input/output), and per-pipeline billing cost in US dollars. The time spent in Reviewer is related to the complexity of models, which explains the increase in TSLib.
}

\label{tab:overall_result}
\centering
\setlength{\tabcolsep}{3.3pt}
\renewcommand{\arraystretch}{1.1}
\fontsize{7.5}{12}\selectfont 
\begin{tabular}{l | c c c c}
\toprule
\textbf{Libraries} & \textbf{Success Rate\tiny{ (\%)}} & \textbf{Time\tiny{ (s)}} & \textbf{{In/Out} Tokens} & \textbf{Cost\tiny{ (US \$)}} \\
\midrule
% PyOD & 77.8\% & 23.8$\pm$4.2 & 3,167$\pm$377 \\
% PyGOD & 75.0\% & 19.5$\pm$3.2 & 3,443$\pm$388 \\
% TSLib & x$\pm$x & x$\pm$x & x$\pm$x \\
PyOD & 100.0 & 24.0 & 3,272/667 & 0.015 \\
PyGOD & 91.1 & 19.6 & 3,143/673 & 0.015 \\
TSLib & 90.0 & 125.2 & 2,680/561 & 0.012\\
% \midrule
% \multicolumn{4}{r}{\multirow{1}{*}{\textbf{Average Cost: }}} \\
\bottomrule
\end{tabular}
\vspace{-14pt}
\end{table}

\vspace{-5pt}
\subsection{Pipeline Generation}
\vspace{-2pt}
We first assess whether \ours can successfully generate runnable pipelines across datasets and models in each supported library. 
We use \textit{GPT-4o}~\citep{gpt4o} to build all agents in our study.
Table~\ref{tab:overall_result} presents the success rate, indicating whether the generated code runs without errors, the average generation time, and the average LLM token usage across different dataset–model pairs.

\ours demonstrates high reliability in producing valid pipelines across modalities, with low latency and manageable cost. 
We provide a complete example run in Appx.~\ref{sec:appx_example_run} for reference.

% \tk{Experiments are running. Need to modify later!!}

\noindent
\textbf{Correction Discussion}. 
The feedback loop between the Code Generator and Reviewer often automatically corrects errors that occur during the initial code generation process. The most frequently fixed issues include missing or incorrectly assigned parameters and incorrect model import names.
For example, when the Generator omits a required argument such as \texttt{n\_features} for DeepSVDD, the Reviewer detects the resulting \texttt{TypeError}, references the correct constructor signature via the Info Miner, and amends the script accordingly.
These correction cases demonstrate the practical benefit of the collaborative agent loop, allowing \ours to recover from common errors and increasing the pipeline success rate without user intervention.

\begin{figure}[t]
\setlength{\abovecaptionskip}{10pt}
\includegraphics[width=0.475\textwidth]{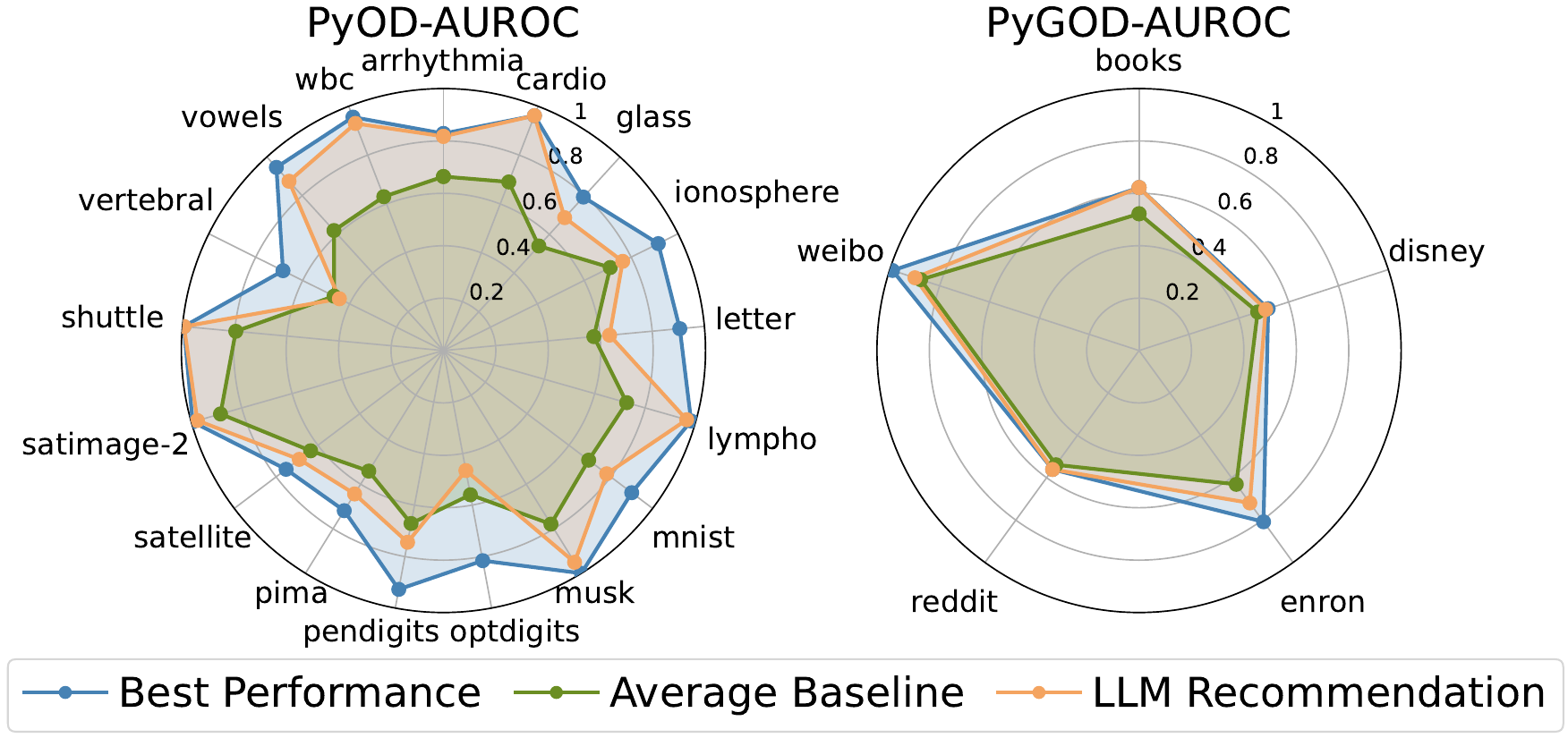}
\vspace{-7pt}
    \caption{Model selection results for PyOD and PyGOD. We display the average AUROC of models recommended by querying the reasoning LLM three times (duplicates allowed). ``Best Performance'' marks the highest performance achieved by any available model for each dataset, while ``Average Baseline'' denotes the mean performance across all available models.
   }
   \label{fig:model_selection_pyod_pygod}
   \vspace{-17pt}
\end{figure}

\noindent
\textbf{Failure Discussion}. 
While \ours demonstrates high overall reliability, a few recurring failure modes remain. Some failures arise from unaddressed internal data constraints. For instance, GAAN in PyGOD expects binary targets for its loss function, but the pipeline sometimes provides values outside the valid range. This highlights the need for improved data validation and type checking within both the Processor and Generator.

% \tk{waiting for one validation.}
Additionally, some errors stem from library inconsistencies or incorrect functions, such as failed imports of DOMINAT in PyGOD, which is therefore excluded from the experiments, or input-size mismatches for Pyraformer in TSLib with certain datasets. 
While these are external, they underscore the need for \ours to integrate version checking and more robust fallback mechanisms.

\vspace{-5pt}
\subsection{Model Selection}
\vspace{-3pt}

% We employ a recent reasoning LLM o4-mini~\citep{gpto4-mini} to select models. Figure~\ref{fig:model_selection_pyod_pygod} illustrates the model selection performance for both PyOD and PyGOD, compared against two reference baselines: (\textit{i}) the \textbf{best} result from any available model, indicating the upper performance limit; and (\textit{ii}) the \textbf{average} performance of all available models, representing random selection.

We employ \textit{o4-mini}~\citep{gpto4-mini} to recommend AD models when the user leaves it unspecified. For each dataset, we query the LLM three times and compute the mean AUROC of selected models. Figure~\ref{fig:model_selection_pyod_pygod} compares the results in PyOD and PyGOD against two baselines: (\textit{i}) the \textbf{best} result from any available model, indicating the upper performance limit; and (\textit{ii}) the \textbf{average} performance of all available models, representing random selection. See more details and results in Appx.~\ref{sec:appx_tslib_model_selection}.

The LLM’s recommendations substantially exceed the average baseline and closely track the best performance in most datasets. This demonstrates that the Selector agent can harness LLM reasoning to choose proper models, simplifying model selection for non-expert users.

\vspace{-5pt}
\subsection{Long-term Memory Efficiency}
\vspace{-2pt}
To quantify the benefit of long-term memory, we compare the Info Miner’s lookup latency and cost when using Web Search versus cached summaries. A typical Web Search takes about 10 seconds, as shown in Table~\ref{tab:long_time_saved}, and costs 0.035 (US \$) per call. In contrast, retrieving the same information from long-term memory is almost instantaneous and incurs no additional cost. This highlights the efficiency of long-term memory.

% We evaluate the efficiency gains from long-term memory by comparing the time spent retrieving algorithm information via Web Search versus using cached summaries. Since cache access is nearly instantaneous, we report only the time for web-based retrieval in Table~\ref{tab:long_time_saved}. This shows that the long-term memory substantially reduces latency.

\begin{table}[t]
% \vspace{-0.3in}
\setlength{\abovecaptionskip}{3pt}
\caption{
% The average latency for the Info Miner to retrieve algorithm metadata via web search for each library is provided. Note that long-term memory lookups are completed instantly and are not included in this data.
Average Web Search latency. Long-term memory lookups complete instantly and are omitted.
}
\label{tab:long_time_saved}
\centering
\setlength{\tabcolsep}{15pt}
\renewcommand{\arraystretch}{1}
\fontsize{8}{12}\selectfont 
\begin{tabular}{l | c c c}
\toprule
\textbf{Libraries} & {PyOD} & {PyGOD} & {TSLib} \\
\midrule
\textbf{Time\tiny{ (s)}} & 10.6 & 12.0 & 10.8 \\
\bottomrule
\end{tabular}

\vspace{-17pt}
\end{table}

\vspace{-3pt}
\section{Conclusion}
\vspace{-4pt}
In this work, we introduced \ours, an LLM-powered multi-agent framework that automates end-to-end AD across multivariate, graph, and time-series data. By decomposing the workflow into specialized agents and coordinating them through short-term and long-term memory, \ours turns natural language instructions into runnable detection pipelines. Our experiments demonstrate high success rates of the system, accurate model recommendations, and substantial reductions in lookup latency and cost via long-term caching. The system is released for further research.

% \noindent
% \textbf{Future Directions.}
% We plan to extend \ours in several directions:
% \begin{itemize}[nosep, itemsep=3pt, leftmargin=*]
%     \item \textbf{Broader Adaption}. Incorporate more data types and libraries.
%     \item \textbf{User Interaction}. Enable users to provide iterative feedback and have agents adapt pipelines.
%     \item \textbf{Online Workspace}. Provide a cloud-based workspace with a pre-configured environment to simplify setup, but ensure data privacy and security at the same time.
%     \item \textbf{Cost-Aware Planning}. Introduce budget constraints so that agents (e.g., Selector and Info Miner) can trade off performance and LLM API costs.
%     \item \textbf{Long-term Sustainability}. Envision a global, community-driven platform where diverse stakeholders collaborate on open-source tools to tackle complex anomaly-detection challenges.
% \end{itemize}

\noindent
\textbf{Future Directions.} We plan to: \textit{(i)} broaden \ours by continually adding new libraries and adapting other data modalities; \textit{(ii)} support conversational interactions so users can iteratively refine pipelines; \textit{(iii)} provide a secure, cloud-based workspace with pre-configured environments to simplify setup; \textit{(iv)} introduce cost-aware planning that balances performance and LLM API budgets; and \textit{(v)} envision a global, community-driven ecosystem where stakeholders collaborate on open-source tools for AD.

\newpage

\section*{Limitations}
Despite its flexibility and automation, \ours has several limitations. The system depends on the accuracy and currency of both the underlying LLMs and external libraries; breaking changes or undocumented features may lead to pipeline failures. Also, not all model or data-specific constraints can be automatically detected, which may result in occasional misconfigurations or runtime errors. Furthermore, \ours has been validated primarily on standard benchmarks, and its effectiveness and robustness for specialized or proprietary datasets need further systematic investigation.

\section*{Ethics Statement}
This work adheres to established ethical standards in both research and software development. All experiments are conducted on public datasets, with no personally identifiable or sensitive information processed or disclosed. 
\ours is under the BSD 2-clause License, ensuring transparency and reproducibility. The system is designed to assist non-expert users in building AD pipelines.
Additionally, ChatGPT was used exclusively to make minor grammatical improvements to the manuscript.

\bibliography{custom}

\newpage
\appendix
\label{sec:appendix}
\section*{Appendix: \ours: A Multi-agent Framework for End-to-end Anomaly Detection}

\section{Related Works}
% \vspace{-4pt}
\label{sec:related}
LLM-based multi-agent systems have emerged as a powerful paradigm for solving complex tasks through role specialization, planning, and tool use~\citep{guo2024survey,li2023camel}. 

These systems have been successfully applied to domains such as software engineering~\citep{liu2024software}, scientific discovery~\citep{liu2024drugagent}, faithfulness evaluation~\citep{koupaee2025faithful}, and social simulations~\citep{li2024political}. 
In the context of AD, Audit-LLM~\citep{song2024audit} targets insider threat detection through multi-agent coordination, and Argos~\citep{gu2025argos} uses LLM agents to generate interpretable anomaly rules for time-series monitoring. While effective, these systems are domain-specific and fixed in scope.

In parallel, several open-source libraries have been developed across different data modalities. 
Popular libraries such as PyOD~\citep{chen2024pyod}, PyGOD~\citep{liu2024pygod}, and TSLib~\citep{tslib} provide strong support for AD on multivariate, graph, and time series data, respectively. 
While each library is effective within its domain, they differ in requirements and design. These inconsistencies make integration across libraries non-trivial.

% \ours brings them together by integrating multiple AD libraries into an LLM-driven multi-agent framework. 
\ours unifies multiple AD libraries within an LLM-driven multi-agent framework.

% \vspace{-3pt}
\section{Experiments Details}
% \vspace{-2pt}
\subsection{Datasets and Models}
% \vspace{-2pt}
\label{sec:appx_dataset_model}
As mentioned in \S~\ref{sec:exp_benchmark}, we adopt datasets and models for each library from corresponding benchmarks.

% \vspace{-3pt}
\subsubsection{PyOD}
Following PyOD 2~\citep{chen2024pyod}, we evaluated \ours on 17 widely used datasets originally from ADBench~\citep{han2022adbench}, including arrhythmia, cardio, glass, ionosphere, letter, lympho, mnist, musk, optdigits, pendigits, pima, satellite, satimage-2, shuttle, vertebral, vowels, and WBC. 
For each dataset, we consider 10 models: ALAD, AnoGAN, AE, AE1SVM, DeepSVDD, DevNet, LUNAR, MO-GAAL, SO-GAAL, and VAE.
See more details in \citet{chen2024pyod}.

% \vspace{-3pt}
\subsubsection{PyGOD}
Following PyGOD~\citep{liu2024pygod}, we evaluated \ours on 5 real datasets originally from BOND~\citep{liu2022bond}, including books, disney, enron, reddit, weibo.
For each dataset, we consider 9 models: AdONE, ANOMALOUS, AnomalyDAE, CONAD, DONE, GAAN, GUIDE, Radar, and SCAN.
See more details in \citet{liu2022bond}.

\begin{table}[t]
\centering
\setlength{\abovecaptionskip}{3pt}
\caption{Detection Performance before and after Optimizer. Better results are highlighted in \textbf{bold}.}
\label{tab:optimizer_results}
\setlength{\tabcolsep}{2.5pt}
\renewcommand{\arraystretch}{1.1}
\fontsize{8}{12}\selectfont 
\begin{tabular}{lcc}
\toprule
\textbf{Models} & \textbf{AUROC(before$\rightarrow$ after)} & \textbf{AUPRC(before$\rightarrow$ after)} \\
\midrule
AE & 0.7875 $\rightarrow$ \textbf{0.8732} & 0.4191 $\rightarrow$ \textbf{0.4959} \\
ALAD        & 0.5861 $\rightarrow$ \textbf{0.6103} & 0.1454 $\rightarrow$ \textbf{0.1624} \\
AnoGAN      & 0.8820 $\rightarrow$ \textbf{0.9438} & 0.6050 $\rightarrow$ \textbf{0.7034} \\
AE1SVM      & 0.9450 $\rightarrow$ \textbf{0.9779} & 0.6748 $\rightarrow$ \textbf{0.8388} \\
DeepSVDD    & 0.9259 $\rightarrow$ \textbf{0.9757} & 0.6370 $\rightarrow$ \textbf{0.8046} \\
DevNet      & 0.0323 $\rightarrow$ 0.0323 & 0.0585 $\rightarrow$ 0.0585 \\
LUNAR       & 0.5254 $\rightarrow$ \textbf{0.7941} & 0.1736 $\rightarrow$ \textbf{0.4462} \\
MO-GAAL     & 0.5300 $\rightarrow$ \textbf{0.6200} & 0.1900 $\rightarrow$ \textbf{0.2000} \\
SO-GAAL     & 0.6687 $\rightarrow$ \textbf{0.7724} & 0.3512 $\rightarrow$ \textbf{0.4283} \\
VAE         & 0.9800 $\rightarrow$ 0.9800 & 0.8300 $\rightarrow$ 0.8300 \\
\bottomrule
\vspace{-20pt}
\end{tabular}
\end{table}

% \vspace{-3pt}
\subsubsection{TSLib}
\citet{tslib_bench} presents a benchmark study for TSLib~\citep{tslib}. Following their approach, we evaluated \ours on 5 real-world datasets from \citet{tslib_bench}, including MSL, PSM, SMAP, SMD, and SWaT.
For each dataset, we consider 10 models: Autoformer, DLinear, ETSformer, FEDformer, Informer, LightTS, Pyraformer, Reformer, TimesNet, and Transformer.
See more details in \citet{tslib_bench}.

% \vspace{-3pt}
\subsection{Use Cases Discussion}
% \vspace{-2pt}
\label{sec:appx_use_case}
Our framework supports two common use cases frequently encountered in academic research and real-world deployments. 

In research or benchmarking settings, users usually have access to a train/test split and ground-truth anomaly labels for the test set. \ours ingests the training data, builds the model, and reports metrics such as AUROC or F1 on the held-out test set if the user enables the Evaluator. Then the Optimizer can further refine hyperparameters by running an inner loop on the training data and passing a possibly better configuration back to the main pipeline before the final evaluation. This mirrors the evaluation protocol adopted by major  AD benchmarks such as ADBench~\citep{han2022adbench}.

In many production scenarios, only one raw, unlabeled dataset is available, and the goal is to identify anomalies directly within this set~\citep{bouman2024unsupervised}. In this case, \ours detects anomalies on the provided data in a single pass; the Evaluator and Optimizer remain inactive unless the user later supplies labels or a separate tuning set.

% \vspace{-3pt}
\subsection{Optimizer Improvement}
% \vspace{-2pt}
\label{sec:appx_optimizer}
To demonstrate the impact of Optimizer, we evaluated it on the dataset ``cardio'' within PyOD. As shown in Table~\ref{tab:optimizer_results}, Optimizer consistently improved detection quality. These results indicate that the Optimizer agent can automatically refine hyperparameters to produce significantly stronger AD pipelines without human intervention.

\subsection{Additional Result of Model Selection}
\label{sec:appx_tslib_model_selection}
\begin{figure}[t]
\setlength{\abovecaptionskip}{10pt}
\includegraphics[width=0.475\textwidth]{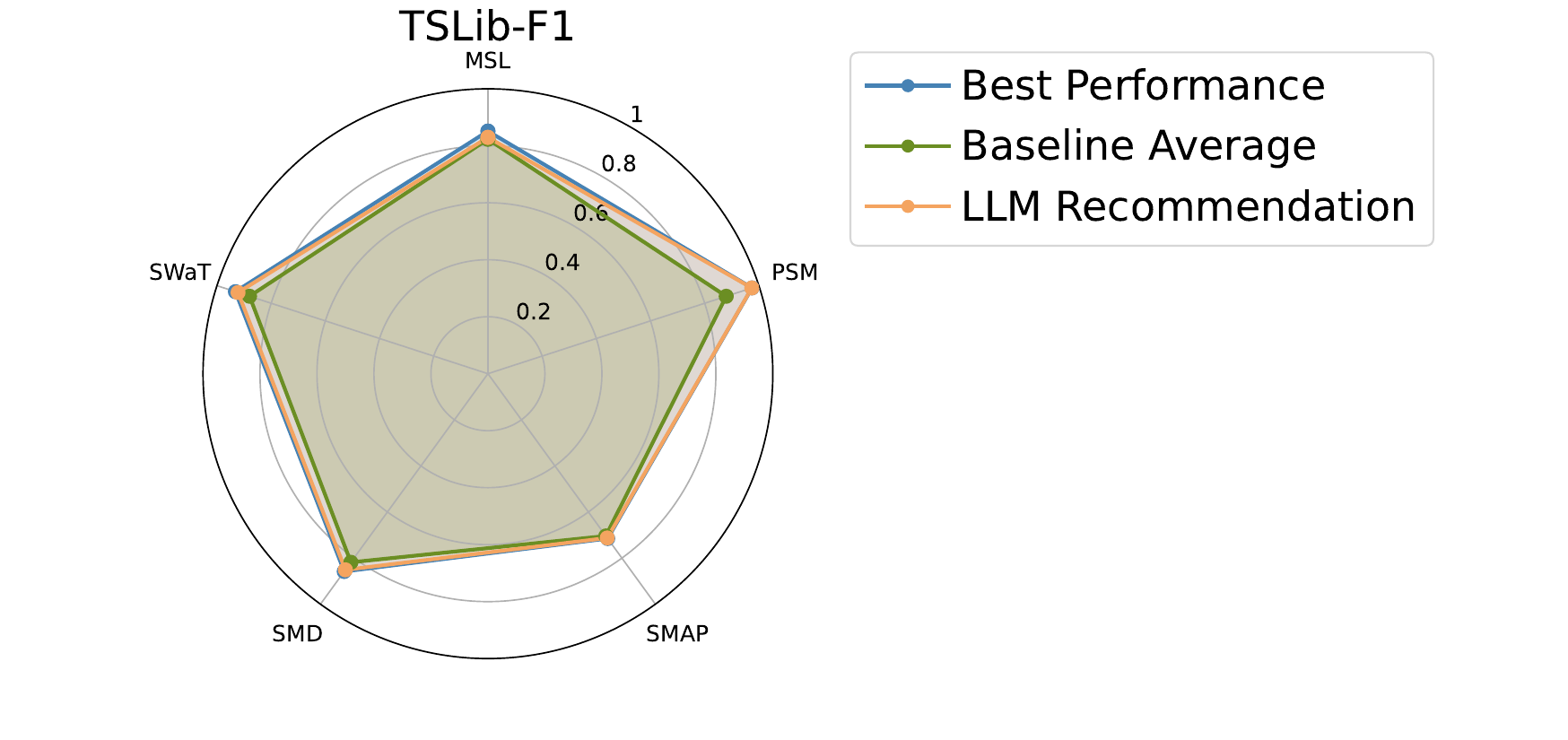}
\vspace{-10pt}
    \caption{Model selection results for TSLib. We display the average F1-score of models recommended by querying the reasoning LLM three times (duplicates allowed). ``Best Performance'' marks the highest performance achieved by any available model for each dataset, while ``Average Baseline'' denotes the mean performance across all available models.
   }
   \label{fig:model_selection_tslib}
   \vspace{-12pt}
\end{figure}

Figure~\ref{fig:model_selection_tslib} shows the model selection results in TSLib. LLM recommendation outperforms the average baseline in all datasets.

\section{Example Run}
\label{sec:appx_example_run}
Table~\ref{tab:example_run} presents an actual session of \ours. 
In this example, a user requests to run VAE on the ``cardio.mat'' dataset via a simple natural language command: ``\textcolor{keygreen}{Run VAE on cardio.mat}.''
The system interprets the user’s intent, processes the data, selects the appropriate library, retrieves model information, and automatically generates a runnable Python script.
This example demonstrates the seamless collaboration between agents in \ours, showing how a single natural language instruction can be transformed into a ready-to-run AD pipeline with minimal user effort.

\begin{table*}[t]
\centering
% \renewcommand{\arraystretch}{0.95}
% \fontsize{10}{12}\selectfont 
% \begin{tabular}{p{0.95\textwidth}}
\begin{tcolorbox} [colback=yellow!10]
=== \text{[}Main\text{]} Starting full pipeline ===\\
=== \text{[}Processor\text{]} Processing user input ===\\
Enter command (e.g., `Run IForest on glass\_train.mat and glass\_test.mat'):\\
User: \textcolor{keygreen}{Run VAE on cardio.mat}\\
Experiment Configuration:\\
\hspace*{1em}Algorithm: [`VAE']\\
\hspace*{1em}Training Dataset: ./data/cardio.mat\\
\hspace*{1em}Testing Dataset: \\
\hspace*{1em}Parameters: {}\\
=== \text{[}Processor\text{]} User input processing complete\\
\textcolor{gray}{-- -- -- -- -- -- -- -- -- -- -- -- -- -- -- -- -- -- -- -- -- -- -- -- -- -- -- -- -- -- -- -- -- -- -- -- -- -- -- -- -- -- -- -- -- -- -- -- -- -- -- -- -- -- }\\
=== \text{[}Selector\text{]} Processing user input ===\\
=== \text{[}Selector\text{]} Selecting package \& algorithm ===\\
Package name: pyod\\
\texttt{\{more content is omitted...\}}\\
=== \text{[}Selector\text{]} Selection complete ===\\
\textcolor{gray}{-- -- -- -- -- -- -- -- -- -- -- -- -- -- -- -- -- -- -- -- -- -- -- -- -- -- -- -- -- -- -- -- -- -- -- -- -- -- -- -- -- -- -- -- -- -- -- -- -- -- -- -- -- -- }\\
=== [Info Miner] Querying documentation for VAE ===\\
% \text{[}Cache Hit\text{]} Using recent cache for VAE\\
The Variational Autoencoder (VAE) in PyOD is \texttt{\{more content is omitted...\}}\\
**Initialization Function (`\_\_init\_\_'):**\\
The `\_\_init\_\_' method initializes the VAE model with \texttt{\{more content is omitted...\}}\\
**Parameters:**\\
- `contamination':\\
\hspace*{1em}- **Type**: float in (0., 0.5)\\
\hspace*{1em}- **Default**: 0.1\\
\hspace*{1em}- **Description**: The proportion of outliers\texttt{\{more content is omitted...\}}\\
- `encoder\_neuron\_list':\\
\texttt{\{more content is omitted...\}}\\
**Attributes:**\\
- `model':\\
\hspace*{1em}- **Type**: torch.nn.Module\\
\hspace*{1em}- **Description**: The underlying VAE model.\\
% - `optimizer':\\
\texttt{\{more content is omitted...\}}\\
**Python Dictionary of `\_\_init\_\_' Parameters with Default Values:**\\
\{\\
\hspace*{1em}`contamination'': 0.1,\\
\hspace*{1em}``encoder\_neuron\_list'': [128, 64, 32],\\
\hspace*{1em}\texttt{\{more content is omitted...\}}\\
\}\\
\text{[}Cache Updated\text{]} Stored new documentation for VAE\\
=== [Info Miner] Documentation retrieved for VAE ===\\
\textcolor{gray}{-- -- -- -- -- -- -- -- -- -- -- -- -- -- -- -- -- -- -- -- -- -- -- -- -- -- -- -- -- -- -- -- -- -- -- -- -- -- -- -- -- -- -- -- -- -- -- -- -- -- -- -- -- -- }\\
=== [Code Generator] Generating code for VAE ===\\
=== [Code Reviewer] Validating for VAE ===\\
=== [Code Reviewer] Validation completed for VAE ===\\
=== [Code Generator] Saved code to ./generated\_scripts/VAE\_cardio.py ===\\
\texttt{\{more content is omitted...\}}

% \bottomrule
% \end{tabular}
\end{tcolorbox}
\vspace{-10pt}
\caption{A real example of \ours. The user provides a single natural language instruction (highlighted in \textcolor{keygreen}{green}), and the system automatically parses the command, retrieves model metadata, and generates an executable Python script. Portions of the printed text are omitted (\texttt{\{more content is omitted...\}}) for brevity.}
\label{tab:example_run}
\end{table*}

% \section{Example Appendix}
% \label{sec:appendix}

% This is an appendix.

\end{document}